\documentclass[11pt,a4paper]{article}
\usepackage{breakurl}
\usepackage[hyperref]{acl2019}
\usepackage{times}
\usepackage{latexsym}
\usepackage{amsmath}
\usepackage{amssymb}
\usepackage{multirow,makecell}
\usepackage{graphicx}
\usepackage[utf8]{inputenc}
\usepackage{booktabs}
\usepackage{pifont}
\usepackage{breqn}
\usepackage{siunitx}
\usepackage{xcolor,colortbl}
\usepackage{tikz}
\usetikzlibrary{shapes,arrows,shadows}
\usepackage{amssymb}
\usepackage{amsmath,bm,times}

\aclfinalcopy 
\def\aclpaperid{1992} 

\pgfdeclarelayer{background}
\pgfdeclarelayer{foreground}
\pgfsetlayers{background,main,foreground}

\tikzstyle{sensor}=[draw, fill=yellow!20, text width=3em, 
    text centered, minimum height=1.5em, rounded corners]
\tikzstyle{output}=[draw, fill=yellow!20, text width=3em, 
    text centered, minimum height=1.5em, rounded corners]
\tikzstyle{ann} = [above, text width=5em, text centered]
\tikzstyle{wa} = [sensor, text width=6em, fill=blue!20, 
    minimum height=6em, rounded corners]
\tikzstyle{sc} = [sensor, text width=13em, fill=blue!20, 
    minimum height=10em, rounded corners]
\tikzstyle{tip}=[draw, fill=blue!20, text width=7em, 
    text centered, minimum height=1.5em, rounded corners]
\tikzstyle{tipnon}=[draw, fill=gray!20, text width=7em, 
    text centered, minimum height=1.5em, rounded corners]

\title{Multilingual Constituency Parsing with Self-Attention and Pre-Training}

\author{Nikita Kitaev \qquad Steven Cao \qquad Dan Klein \\
  Computer Science Division \\
  University of California, Berkeley \\
  {\tt \{kitaev,stevencao,klein\}@berkeley.edu}}

\date{}

\begin{document}
\maketitle
\begin{abstract}
We show that constituency parsing benefits from unsupervised pre-training across a variety of languages and a range of pre-training conditions. We first compare the benefits of no pre-training, fastText~\citep{bojanowski2017enriching,mikolov2018advances}, ELMo~\citep{peters_deep_2018}, and BERT~\citep{devlin_bert:2018} for English and find that BERT outperforms ELMo, in large part due to increased model capacity, whereas ELMo in turn outperforms the non-contextual fastText embeddings. We also find that pre-training is beneficial across all 11 languages tested; however, large model sizes (more than 100 million parameters) make it computationally expensive to train separate models for each language. To address this shortcoming, we show that joint multilingual pre-training and fine-tuning allows sharing all but a small number of parameters between ten languages in the final model. The 10x reduction in model size compared to fine-tuning one model per language causes only a 3.2\% relative error increase in aggregate. We further explore the idea of joint fine-tuning and show that it gives low-resource languages a way to benefit from the larger datasets of other languages. Finally, we demonstrate new state-of-the-art results for 11 languages, including English (95.8~F1) and Chinese (91.8 F1).
\end{abstract}

\section{Introduction}
\label{sec:intro}

There has recently been rapid progress in developing contextual word representations that improve accuracy across a range of natural language tasks~\cite{peters_deep_2018,howard_2018_ulmfit,radford2018improving,devlin_bert:2018}. While we have shown in previous work~\citep{kitaev_2018_self_attentive} that such representations are beneficial for constituency parsing, our earlier results only consider the LSTM-based ELMo representations~\cite{peters_deep_2018}, and only for the English language. In this work, we study a broader range of pre-training conditions and experiment over a variety of languages, both jointly and individually.

First, we consider the impact on parsing of using
different methods for pre-training initial network layers on a large collection of un-annotated text.
Here, we see that pre-training provides benefits for all languages evaluated, and that BERT~\cite{devlin_bert:2018} outperforms ELMo, which in turn outperforms fastText~\citep{bojanowski2017enriching,mikolov2018advances}, which performs slightly better than the non pre-trained baselines. Pre-training with a larger model capacity typically leads to higher parsing accuracies.

Second, we consider various schemes for the parser fine-tuning that is required after pre-training. While BERT itself can be pre-trained jointly on many languages, successfully applying it, e.g.\ to parsing, requires task-specific adaptation via fine-tuning \cite{devlin_bert:2018}.  Therefore, the obvious approach to parsing ten languages is to fine-tune ten times, producing ten variants of the parameter-heavy BERT layers.  In this work, we compare this naive independent approach to a joint fine-tuning method where a single copy of fine-tuned BERT parameters is shared across all ten languages.  Since only a small output-specific fragment of the network is unique to each task, the model is 10x smaller while losing an average of only 0.28 F1.

Although, in general, jointly training multilingual parsers mostly provides a more compact model, it does in some cases improve accuracy as well. To investigate when joint training is helpful, we also perform paired fine-tuning on all pairs of languages and examine which pairs lead to the largest increase in accuracy. We find that larger treebanks function better as auxiliary tasks and that only smaller treebanks see a benefit from joint training. These results suggest that this manner of joint training can be used to provide support for many languages in a resource-efficient manner, but does not exhibit substantial cross-lingual generalization except when labeled data is limited.
Our parser code and trained models for eleven languages are publicly available.\footnote{\href{https://github.com/nikitakit/self-attentive-parser}{https://github.com/nikitakit/self-attentive-parser}}

\section{Model}

Our parsing model is based on the architecture described in \citet{kitaev_2018_self_attentive}, which is state of the art for multiple languages, including English.
A constituency tree $T$ is represented as a set of labeled spans,
\[
T = \left\{(i_t, j_t, l_t) : t=1,\ldots,\left|T\right|\right\}
\]
where the $t^{\text{th}}$ span begins at position $i_t$, ends at position $j_t$, and has label $l_t$. The parser assigns a score $s(T)$ to each tree, which decomposes as
\[
s(T) = \sum_{(i,j,l) \in T} s(i,j,l)
\]
The per-span scores $s(i,j,l)$ are produced by a neural network. This network accepts as input a sequence of vectors corresponding to words in a sentence and transforms these representations using one or more self-attention layers. For each span $(i,j)$ in the sentence, a hidden vector $v_{i,j}$ is constructed by subtracting the representations associated with the start and end of the span. An MLP span classifier, consisting of two fully-connected layers with one ReLU nonlinearity, assigns labeling scores $s(i,j,\cdot)$ to the span. Finally, the the highest scoring valid tree \[\hat{T} = \arg\max_T s(T)\] can be found efficiently using a variant of the CKY algorithm. For more details, see \citet{kitaev_2018_self_attentive}.

We incorporate BERT by computing token representations from the last layer of a BERT model, applying a learned projection matrix, and then passing them as input to the parser.
BERT associates vectors to sub-word units based on WordPiece tokenization \citep{wu_googles_2016}, from which we extract word-aligned representations by only retaining the BERT vectors corresponding to the last sub-word unit for each word in the sentence.
We briefly experimented with other alternatives, such as using only the first sub-word instead, but did not find that this choice had a substantial effect on English parsing accuracy.

The fact that additional layers are applied to the output of BERT -- which itself uses a self-attentive architecture -- may at first seem redundant, but there are important differences between these two portions of the architecture. The extra layers on top of BERT use word-based tokenization instead of sub-words, apply the factored version of self-attention proposed in \citet{kitaev_2018_self_attentive}, and are randomly-initialized instead of being pre-trained. We found that passing the (projected) BERT vectors directly to the MLP span classifier hurts parsing accuracies.

We train our parser with a learning rate of \num{5e-5} and batch size 32, where BERT parameters are fine-tuned as part of training. We use two additional self-attention layers following BERT. All other hyperparameters are unchanged from \citet{kitaev_2018_self_attentive} and \citet{devlin_bert:2018}.

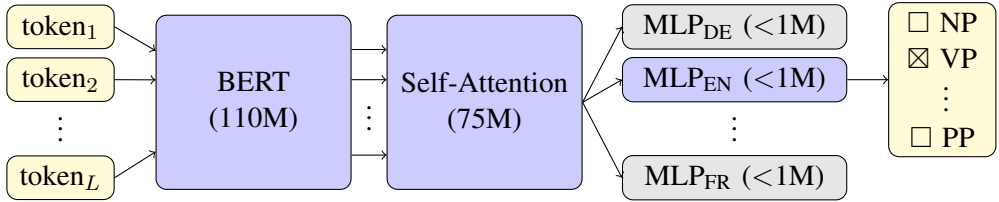
\begin{figure*}
\centering
\begin{tikzpicture}
    \node (bert) [wa]  {BERT \\(110M)};
    \path (bert.west)+(-1.25,1.0) node (asr1) [sensor] {token$_1$};
    \path (bert.west)+(-1.25,0.3) node (asr2)[sensor] {token$_2$};
    \path (bert.west)+(-1.25,-0.65) node (dots)[ann] {$\vdots$}; 
    \path (bert.west)+(-1.25,-1.0) node (asr3)[sensor] {token$_L$};    
   
    \path (bert.east)+(1.75,0) node (attn) [wa] {Self-Attention \\ (75M)};
    
    \path (attn.east)+(2,1.0) node (tip1) [tipnon] {MLP$_{\text{DE}}$ ($<$1M)};
    \path (attn.east)+(2,0.3) node (tip2)[tip] {MLP$_{\text{EN}}$ ($<$1M)};
    \path (attn.east)+(2,-0.65) node (dots2)[ann] {$\vdots$}; 
    \path (attn.east)+(2,-1.0) node (tip3)[tipnon] {MLP$_{\text{FR}}$ ($<$1M)};  
    
    \path (tip2.east)+(1.25,0) node (out)[output] {$\square$ NP \\$\boxtimes$ VP \\ $\hskip 0.1cm \vdots$ \\$\square$ PP}; 

    \path [draw, ->] (asr1.east) -- node [above] {} 
        (bert.152) ;
    \path [draw, ->] (asr2.east) -- node [above] {} 
        (bert.167);
    \path [draw, ->] (asr3.east) -- node [above] {} 
        (bert.207);
    \path [draw, ->] ([yshift=0.3cm]bert.east) -- node [below] {$\vdots$} 
        ([yshift=0.3cm]attn.west);
    \path [draw, ->] ([yshift=0.7cm]bert.east) -- node [above] {} 
        ([yshift=0.7cm]attn.west);
    \path [draw, ->] ([yshift=-0.7cm]bert.east) -- node [above] {} 
        ([yshift=-0.7cm]attn.west);
    \path [draw, ->] (attn.east) -- node [above] {} 
        (tip1.180);
    \path [draw, ->] (attn.east) -- node [above] {} 
        (tip2.180);
    \path [draw, ->] (attn.east) -- node [above] {} 
        (tip3.180);
     \path [draw, ->] (tip2.east) -- node [above] {}
        (out.180);
\end{tikzpicture}
\caption{\label{figure:arch}
The architecture of the multilingual model, with components labeled by the number of parameters.
}
\end{figure*}

\section{Comparison of Pre-Training Methods}
\label{sec:comparison-english}

\begin{table}[t]
\begin{center}
\resizebox{1.0\linewidth}{!}{
\begin{tabular}{@{}lllc@{}}
\toprule
Method & Pre-trained on & Params & F1  \\
\midrule
No pre-training & --  & \phantom{0}26M & 93.61$^a$ \\
FastText& English & 626M & 93.72\phantom{$^a$} \\
ELMo& English & 107M & 95.21$^a$ \\
BERT\textsubscript{BASE} (uncased)& Chinese & 110M  & 93.57\phantom{$^a$}\\
BERT\textsubscript{BASE} (cased) & 104 languages & 185M & 94.97\phantom{$^a$} \\
BERT\textsubscript{BASE} (uncased)& English & 117M & 95.32\phantom{$^a$} \\
BERT\textsubscript{BASE} (cased) & English & 116M & 95.24\phantom{$^a$} \\
BERT\textsubscript{LARGE} (uncased) & English & 343M & 95.66\phantom{$^a$} \\
BERT\textsubscript{LARGE} (cased) & English & 341M & 95.70\phantom{$^a$} \\
\midrule
\multicolumn{2}{@{}l}{Ensemble (final 4 models above)} &916M & \textbf{95.87}\phantom{$^a$} \\
\bottomrule
\end{tabular}
} 
\end{center}
\caption{\label{table:wsj-dev} Comparison of parsing accuracy on the WSJ development set when using different word representations. $^a$\citet{kitaev_2018_self_attentive}}
\end{table}

In this section, we compare using BERT, ELMo, fastText, and training a parser from scratch on treebank data alone.
Our comparison of the different methods for English is shown in Table~\ref{table:wsj-dev}.
BERT\textsubscript{BASE}
($\sim$115M parameters)
performs comparably or slightly better than ELMo ($\sim$107M parameters; 95.32 vs.\ 95.21 F1), while
BERT\textsubscript{LARGE}
($\sim$340M parameters)
leads to better parsing accuracy (95.70 F1). Furthermore, both pre-trained contextual embeddings significantly outperform fastText, which performs slightly better than no pre-training (93.72 vs.\ 93.61 F1). These results show that both the LSTM-based architecture of ELMo and the self-attentive architecture of BERT are viable for parsing, and that pre-training benefits from having a high model capacity. We did not observe a sizable difference between an ``uncased'' version of BERT that converts all text to lowercase and a ``cased'' version of that retains case information.

\begin{table*}[h]
\centering
\resizebox{1.0\linewidth}{!}{
\begin{tabular}{@{}lcccccccccccc@{}}
\toprule
  &Arabic &Basque &English &French &German &Hebrew &Hungarian &Korean &Polish &Swedish &Avg& Params \\
  \midrule
  No pre-training$^a$ &85.61&89.71&93.55&84.06&87.69&90.35&92.69&86.59&93.69&84.45&88.32 & \phantom{1,}355M\\ 
  One model per language (this work) &\textbf{87.97} &\textbf{91.63} &\textbf{94.91} &\textbf{87.42} &\textbf{90.20} &\textbf{92.99} &\textbf{94.90} &88.80 &\textbf{96.36} &88.86 &\textbf{91.40} & 1,851M\\  
  Joint multilingual model (this work) &87.44 &90.70 &94.63 &87.35 &88.40 &92.95 &94.60 &\textbf{88.96} &96.26 &\textbf{89.94} &91.12 & \phantom{1,}189M\\
  \midrule
  Relative $\Delta$Error vs. monolingual &+4.2\%\rlap{*} &+10.0\%\rlap{*}&+5.2\%\rlap{*}&+0.6\%&+15.5\%\rlap{*}&+0.6\%&+5.6\%\rlap{*}&-1.5\%&+2.7\%&-10.7\%\rlap{*}& +3.2\%\rlap{*}\\
  \bottomrule
\end{tabular}
}
\caption{\label{table:multi}Results of monolingual and multilingual training on the SPMRL and WSJ test splits using the version of BERT pre-trained on 104 languages. {\iffalse Other than for English, F1 scores are calculated using the version of \texttt{evalb} distributed with the SMPRL shared task.\fi } In the last row, starred differences are significant at the $p < 0.05$ level using a bootstrap test; see~\citet{berg-kirkpatrick-etal-2012-empirical}. $^a$\citet{kitaev_2018_self_attentive}
} 
\end{table*}

\begin{table*}[h]
\resizebox{1.0\linewidth}{!}{
\begin{tabular}{@{}lccccccccccccc@{}}
\toprule
  \textbf{Auxiliary Language} &Arabic &Basque &English &French &German &Hebrew &Hungarian &Korean &Polish &Swedish &Average &Best &Best Aux. \\
  \midrule
  \textbf{\# train sentences} &15,762 & 7,577 & 39,831 & 14,759 & 40,472 & 5,000 & 8,146 & 23,010 & 6,578 & 5,000 \\
  \midrule
  \textbf{Language Tested} \\ \addlinespace
Arabic& \cellcolor{yellow!25}\phantom{-}0\phantom{.00}&\cellcolor{red!25}-0.38&\cellcolor{orange!25}-0.20&\cellcolor{orange!25}-0.27&\cellcolor{orange!25}-0.26&\cellcolor{yellow!25}-0.14&\cellcolor{orange!25}-0.29&\cellcolor{yellow!25}-0.13&\cellcolor{red!25}-0.31&\cellcolor{red!25}-0.33& -0.23& \textbf{+0\phantom{.00*}}&None \\
Basque&\cellcolor{red!25}-0.47&\cellcolor{yellow!25}\phantom{-}0\phantom{.00}&\cellcolor{yellow!25}-0.06&\cellcolor{orange!25}-0.26& \cellcolor{yellow!25}\phantom{-}0.04&\cellcolor{orange!25}-0.22&\cellcolor{orange!25}-0.27&\cellcolor{red!25}-0.41&\cellcolor{red!25}-0.49& \cellcolor{red!25}-0.34& -0.25& \textbf{+0.04\phantom{*}}&German\\
English& \cellcolor{orange!25}-0.18& \cellcolor{yellow!25}-0.04& \cellcolor{yellow!25}\phantom{-}0\phantom{.00}& \cellcolor{yellow!25}-0.02& \cellcolor{yellow!25}-0.03& \cellcolor{yellow!25}-0.07& \cellcolor{yellow!25}-0.09& \cellcolor{yellow!25}\phantom{-}0.05& \cellcolor{yellow!25}\phantom{-}0.10& \cellcolor{yellow!25}-0.05& -0.03&  \textbf{+0.10\phantom{*}}&Polish\\
French& \cellcolor{green!25}\phantom{-}0.42& \cellcolor{yellow!25}\phantom{-}0.01& \cellcolor{lime!33}\phantom{-}0.28& \cellcolor{yellow!25}\phantom{-}0\phantom{.00}& \cellcolor{green!25}\phantom{-}0.40& \cellcolor{yellow!25}-0.14& \cellcolor{yellow!25}\phantom{-}0.04& \cellcolor{lime!33}\phantom{-}0.27& \cellcolor{lime!33}\phantom{-}0.29& \cellcolor{yellow!25}-0.10& \phantom{-}0.15& \textbf{+0.42}*& Arabic\\
German& \cellcolor{red!25}-0.38& \cellcolor{orange!25}-0.20& \cellcolor{yellow!25}\phantom{-}0.03& \cellcolor{red!25}-0.45& \cellcolor{yellow!25}\phantom{-}0\phantom{.00}& \cellcolor{yellow!25}-0.13& \cellcolor{yellow!25}-0.15& \cellcolor{yellow!25}-0.13& \cellcolor{orange!25}-0.21& \cellcolor{orange!25}-0.26& -0.19& \textbf{+0.03\phantom{*}}& English\\
Hebrew& \cellcolor{yellow!25}\phantom{-}0.13& \cellcolor{yellow!25}\phantom{-}0.05& \cellcolor{orange!25}-0.27& \cellcolor{orange!25}-0.17& \cellcolor{yellow!25}-0.11& \cellcolor{yellow!25}\phantom{-}0\phantom{.00}& \cellcolor{yellow!25}-0.09& \cellcolor{orange!25}-0.19& \cellcolor{orange!25}-0.30& \cellcolor{red!25}-0.35& -0.13& \textbf{+0.13\phantom{*}}&Arabic\\
Hungarian& \cellcolor{yellow!25}-0.14& \cellcolor{red!25}-0.43& \cellcolor{orange!25}-0.29& \cellcolor{red!25}-0.38& \cellcolor{yellow!25}-0.11& \cellcolor{red!25}-0.39& \cellcolor{yellow!25}\phantom{-}0\phantom{.00}& \cellcolor{orange!25}-0.17& \cellcolor{orange!25}-0.28& \cellcolor{red!25}-0.32& -0.25& \textbf{+0\phantom{.00*}}& None\\
Korean& \cellcolor{orange!25}-0.24& \cellcolor{orange!25}-0.25& \cellcolor{lime!33}\phantom{-}0.16& \cellcolor{orange!25}-0.27& \cellcolor{yellow!25}-0.11& \cellcolor{yellow!25}-0.01& \cellcolor{yellow!25}\phantom{-}0\phantom{.00}& \cellcolor{yellow!25}\phantom{-}0\phantom{.00}& \cellcolor{yellow!25}-0.07& \cellcolor{orange!25}-0.17& -0.10&  \textbf{+0.16\phantom{*}}& English\\
Polish& \cellcolor{lime!33}\phantom{-}0.25& \cellcolor{lime!33}\phantom{-}0.15& \cellcolor{lime!33}\phantom{-}0.20& \cellcolor{lime!33}\phantom{-}0.24& \cellcolor{lime!33}\phantom{-}0.24& \cellcolor{lime!33}\phantom{-}0.21& \cellcolor{yellow!25}\phantom{-}0.14& \cellcolor{lime!33}\phantom{-}0.20& \cellcolor{yellow!25}\phantom{-}0\phantom{.00}& \cellcolor{yellow!25}\phantom{-}0.12& \phantom{-}0.18& \textbf{+0.25}*& Arabic\\
Swedish& \cellcolor{lime!33}\phantom{-}0.17& \cellcolor{yellow!25}-0.08& \cellcolor{green!25}\phantom{-}0.38& \cellcolor{green!25}\phantom{-}0.54& \cellcolor{green!25}\phantom{-}0.53& \cellcolor{yellow!25}-0.11& \cellcolor{green!25}\phantom{-}0.59& \cellcolor{green!25}\phantom{-}0.78& \cellcolor{orange!25}-0.17& \cellcolor{yellow!25}\phantom{-}0\phantom{.00}& \phantom{-}0.26& \textbf{+0.78}*& Korean\\
\midrule
Average& -0.04 & -0.12 & \phantom{-}0.02 & -0.10 & \phantom{-}0.06 & -0.10 & -0.01 & \phantom{-}0.03 & -0.14 & -0.18\\
  \bottomrule
\end{tabular}
}
\caption{\label{table:pairs} Change in development set F1 score due to paired vs. individual fine-tuning. In the ``Best'' column, starred results are significant at the $p < 0.05$ level. On average, the three largest treebanks (German, English, Korean) function the best as auxiliaries. Also, the three languages benefitting most from paired training (Swedish, French, Polish) function poorly as auxiliaries.}
\end{table*}

We also evaluate an ensemble of four English BERT-based parsers, where the models are combined by averaging their span label scores:
\[
s_{\text{ensemble}}(i,j,l) = \frac{1}{4} \sum_{n=1}^4 s_n(i,j,l)
\]
The resulting accuracy increase with respect to the best single model (95.87 F1 vs.\ 95.66 F1) reflects not only randomness during fine-tuning, but also variations between different versions of BERT. When combined with the observation that BERT\textsubscript{LARGE} outperforms BERT\textsubscript{BASE}, the ensemble results suggest that empirical gains from pre-training have not yet plateaued as a function of computational resources and model size.

Next, we compare pre-training on monolingual data to pre-training on data that includes a variety of languages. We find that pre-training on only English outperforms multilingual pre-training given the same model capacity, but the decrease in accuracy is less than 0.3 F1 (95.24 vs.\ 94.97 F1). This is a promising result because it supports the idea of parameter sharing as a way to provide support for many languages in a resource-efficient manner, which we examine further in Section~\ref{sec:multilingual}.

To further examine the effects of pre-training on disparate languages, we consider the extreme case of training an English parser using a version of BERT that was pre-trained on the Chinese Wikipedia.
Neither the pre-training data nor the subword vocabulary used are a good fit for the target task.
However, English words (e.g.\ proper names) occur in the Chinese Wikipedia data with sufficient frequency that the model can losslessly represent English text: all English letters are included in its subword vocabulary, so in the worst case it will decompose an English word into its individual letters. We found that this model achieves performance comparable to our earlier parser~\citep{kitaev_2018_self_attentive} trained on treebank data alone (93.57 vs.\ 93.61 F1). These results suggest that even when the pre-training data is a highly imperfect fit for the target application, fine-tuning can still produce results better than or comparable to purely supervised training with randomly-initialized parameters.\footnote{We also attempted to use a randomly-initialized BERT model, but the resulting parser did not train effectively within the range of hyperparameters we tried. Note that the original BERT models were trained on significantly more powerful hardware and for a longer period of time than any of the experiments we report in this paper.}

\section{Multilingual Model}
\label{sec:multilingual}

We next evaluate how well self-attention and pre-training work cross-linguistically; for this purpose we consider ten languages: English and the nine languages represented in the SPMRL 2013/2014 shared tasks \citep{seddah_overview_2013}.

\begin{table*}
\centering
\resizebox{1.0\linewidth}{!}{
\begin{tabular}{@{}lcccccccccc@{}}
\toprule
  &Arabic &Basque &French &German &Hebrew &Hungarian &Korean &Polish &Swedish &Avg \\
  \midrule
  \citet{bjorkelund_ims-wroclaw-szeged-cis_2014} &
81.32\rlap{$^a$}&88.24&82.53&81.66&89.80&91.72&83.81&90.50&85.50&86.12 \\
  \citet{coavoux_multilingual_2017} &
82.92\rlap{$^b$}&88.81&82.49&85.34&89.87&92.34&86.04&93.64&84.0\phantom{0}&87.27\\
  \citet{kitaev_2018_self_attentive} &85.61\rlap{$^c$}&89.71\rlap{$^c$}&84.06&87.69&90.35&92.69&86.59\rlap{$^c$}&93.69\rlap{$^c$}&84.45&88.32\\
  \addlinespace
  This work (joint multilingual model) &87.44 &90.70 & 87.35 &88.40 &92.95 &94.60 &\textbf{88.96} &96.26 &\textbf{89.94} & 90.73\\ 
$\Delta$ vs. best previous&+1.83&+0.99&+3.29&+0.71&+2.60&+1.91&\textbf{+2.37}&+2.57&\textbf{+4.44} \\ 
\addlinespace
  This work (one model per language) &\textbf{87.97}&\textbf{91.63}&\textbf{87.42}&\textbf{90.20}&\textbf{92.99}&\textbf{94.90}&88.80&\textbf{96.36}&88.86&\textbf{91.01}\\
  $\Delta$ vs. best previous &\textbf{+2.36}&\textbf{+1.92}&\textbf{+3.36}&\textbf{+2.51}&\textbf{+2.64}&\textbf{+2.21}&+2.21&\textbf{+2.67}&+3.36 \\
  \bottomrule
\end{tabular}
}
\caption{\label{table:spmrl}Results on the testing splits of the SPMRL dataset. All values are F1 scores calculated using the version of \texttt{evalb} distributed with the shared task. $^a$\citet{bjorkelund_re_2013} $^b$Uses character LSTM, whereas other results from \citet{coavoux_multilingual_2017} use predicted part-of-speech tags. $^c$Does not use word embeddings, unlike other results from \citet{kitaev_2018_self_attentive}.
} 
\end{table*}

\begin{table}[h]
\begin{center}
\resizebox{1.0\linewidth}{!}{
\begin{tabular}{@{}lccc@{}}
\toprule
& LR & LP & F1  \\
\midrule
\citet{dyer_recurrent_2016} & -- & -- & 93.3\phantom{0} \\
\citet{choe_parsing_2016}\hspace{-1em} & -- & -- & 93.8\phantom{0} \\
\citet{liu_in_order_2017} & -- & -- & 94.2\phantom{0} \\
\citet{fried_improving_2017} & -- & -- & 94.66 \\
\citet{joshi_2018_extending} & 93.8\phantom{0} & 94.8\phantom{0} & 94.3\phantom{0} \\
\citet{kitaev_2018_self_attentive} & 94.85 & 95.40 & 95.13 \\
This work (single model) & 95.46 & 95.73 & 95.59 \\
This work (ensemble of 4) & \textbf{95.51} & \textbf{96.03} & \textbf{95.77} \\
\bottomrule
\end{tabular}
}
\end{center}
\caption{\label{table:wsj-test} Comparison of F1 scores on the WSJ test set.}
\end{table}

Our findings from the previous section show that pre-training continues to benefit from larger model sizes when data is abundant. However, as models grow, it is not scalable to conduct separate pre-training and fine-tuning for all languages. This shortcoming can be partially overcome by pre-training BERT on multiple languages, as suggested by the effectiveness of the English parser fine-tuned from multilingual BERT (see Table~\ref{table:wsj-dev}). Nevertheless, this straightforward approach also faces scalability challenges because it requires training an independent parser for each language, which results in over $1.8$ billion parameters for ten languages. Therefore, we consider a single parser with parameters shared across languages and fine-tuned jointly. The joint parser uses the same BERT model and self-attention layers for all ten languages but contains one MLP span classifier per language to accommodate the different tree labels (see Figure~\ref{figure:arch}). The MLP layers contain 250K-850K parameters, depending on the type of syntactic annotation adopted for the language, which is less than 0.5\% of the total parameters. Therefore, this joint training entails a 10x reduction in model size.

During joint fine-tuning, each batch contains sentences from every language. Each sentence passes through the shared layers and then through the MLP span classifier corresponding to its language.
To reduce over-representation of languages with large training sets, we follow \citet{devlin_github_2018} and determine the sampling proportions through exponential smoothing:
if a language is some fraction $f$ of the joint training set, the probability of sampling examples from that language is proportional to $f^a$ for some $a$.
We use the same hyperparameters as in monolingual training but increase the batch size to 256 to account for the increase in the number of languages, and we use $a = 0.7$ as in \citet{devlin_github_2018}. The individually fine-tuned parsers also use the same hyperparameters, but without the increase in batch size.

Table~\ref{table:multi} presents a comparison of different parsing approaches across a set of ten languages. Our joint multilingual model outperforms treebank-only models~\citep{kitaev_2018_self_attentive} for each of the languages (88.32 vs 91.12 average F1). We also compare joint and individual fine-tuning. The multilingual model on average degrades performance only slightly (91.12 vs. 91.40 F1) despite the sharp model size reduction, and in fact performs better for Swedish.

We hypothesize that the gains/losses in accuracy for different languages stem from two competing effects: the multilingual model has access to more data, but there are now multiple objective functions competing over the same parameters. To examine language compatibility, we also train a bilingual model for each language pair and compare it to the corresponding monolingual model (see Table~\ref{table:pairs}). From this experiment, we see that the best language pairs often do not correspond to any known linguistic groupings, suggesting that compatibility of objective functions is influenced more by other factors such as treebank labeling convention. In addition, we see that on average, the three languages with the largest training sets (English, German, Korean) function well as auxiliaries. Furthermore, the three languages that gain the most from paired training (Swedish, French, Polish) have smaller datasets and function poorly as auxiliaries. These results suggest that joint training not only drastically reduces model size, but also gives languages with small datasets a way to benefit from the large datasets of other languages.

\begin{table}
\begin{center}
\resizebox{1.0\linewidth}{!}{
\begin{tabular}{@{}lccc@{}}
\toprule
& LR & LP & F1  \\
\midrule
\citet{fried_2018_policy_gradient} & -- & -- & 87.0\phantom{0}  \\
\citet{teng_2018_two_local} & 87.1\phantom{0} & 87.5\phantom{0}  & 87.3\phantom{0}  \\
This work & \textbf{91.55} & \textbf{91.96} & \textbf{91.75} \\
\bottomrule
\end{tabular}
}
\end{center}
\caption{\label{table:ctb-test} Comparison of F1 scores on the Chinese Treebank 5.1 test set.}
\end{table}

\section{Results}
\label{sec:results}

We train and evaluate our parsers on treebanks for eleven languages: the nine languages represented in the SPMRL 2013/2014 shared tasks \citep{seddah_overview_2013} (see Table~\ref{table:spmrl}), English (see Table~\ref{table:wsj-test}), and Chinese (see Table~\ref{table:ctb-test}). The English and Chinese parsers use fully monolingual training, while the remaining parsers incorporate a version of BERT pre-trained jointly on 104 languages. For each of these languages, we obtain a higher F1 score than any past systems we are aware of.

In the case of SPRML, both our single multilingual model and our individual monolingual models achieve higher parsing accuracies than previous systems (none of which made use of pre-trained contextual word representations). This result shows that pre-training is beneficial even when model parameters are shared heavily across languages.

\section{Conclusion}

The remarkable effectiveness of unsupervised pre-training of vector representations of language suggests that future advances in this area can continue improving the ability of machine learning methods to model syntax (as well as other aspects of language).
As pre-trained models become increasingly higher-capacity, joint multilingual training is a promising approach to scalably providing NLP systems for a large set of languages.

\section*{Acknowledgments}

This research was supported by DARPA through the XAI program. This work used the Savio computational cluster provided by the Berkeley Research Computing program at the University of California, Berkeley. 

\bibliography{acl2019}
\bibliographystyle{acl_natbib}

\end{document}